\pdfoutput=1
\documentclass[11pt]{article}

\usepackage[]{ACL2023}

\usepackage{times}
\usepackage{latexsym}

\usepackage[T1]{fontenc}
\usepackage[utf8]{inputenc}

\usepackage{microtype}

\usepackage{inconsolata}
\usepackage{CJK}
\usepackage{tabularx}
\usepackage{graphicx}
\usepackage{booktabs}
\usepackage{float}
\usepackage{color}
\usepackage{arydshln}
\usepackage{bm}
\usepackage{amssymb}
\usepackage{multirow}
\usepackage{multicol}
\usepackage{bbding}
\usepackage{arydshln}

\newcommand{\tabincell}[2]{\begin{tabular}{@{}#1@{}}#2\end{tabular}}
\definecolor{mygreen}{RGB}{46,139,87}
\definecolor{myblue}{RGB}{30,144,255}
\definecolor{myred}{RGB}{238,0,0}

\title{\textsc{VCSum}: A Versatile Chinese Meeting Summarization Dataset}

\author{Han Wu$^{1,2}$, Mingjie Zhan$^{3,}$$^\dagger$, Haochen Tan$^{1,2,}$$^\dagger$, Zhaohui Hou$^{3,}$$^\dagger$, Ding Liang$^{3}$, Linqi Song$^{1,2}$\\
$^{1}$ Department of Computer Science, City University of Hong Kong\\
$^{2}$ City University of Hong Kong Shenzhen Research Institute\\
$^{3}$ SenseTime Research\\
\texttt{\{hanwu32-c,haochetan-2\}@my.cityu.edu.hk}\\
\texttt{\{zhanmingjie,houzhaohui,liangding\}@sensetime.com}\\
\texttt{linqi.song@cityu.edu.hk}
}

\begin{document}
\maketitle
\begin{CJK*}{UTF8}{gbsn}
\begin{abstract}
Compared to news and chat summarization, the development of meeting summarization is hugely decelerated by the limited data. To this end, we introduce a versatile Chinese meeting summarization dataset, dubbed \textsc{VCSum}, consisting of 239 real-life meetings, with a total duration of over 230 hours. We claim our dataset is \textit{versatile} because we provide the annotations of topic segmentation, headlines, segmentation summaries, overall meeting summaries, and salient sentences for each meeting transcript. As such, the dataset can adapt to various summarization tasks or methods, including segmentation-based summarization, multi-granularity summarization and retrieval-then-generate summarization. Our analysis confirms the effectiveness and robustness  of \textsc{VCSum}. We also provide a set of benchmark models regarding different downstream summarization tasks on \textsc{VCSum} to facilitate further research. The dataset and code will be released at \url{https://github.com/hahahawu/VCSum}.
\end{abstract}

\section{Introduction}
{
\let\thefootnote\relax\footnotetext{
$^\dagger$Equal contribution.}
}
Meeting summarization \citep{janin2003icsi, mccowan2005ami, zhong2021qmsum} is the task of distilling the meeting transcript into a concise and readable summary that contains the most salient parts of a meeting. The summary can help the participants or absentees to quickly grape the highlight points. Therefore, a set of models have been proposed to comprehensively and succinctly summarize the content of a meeting \citep{zhu2020hierarchical, feng2021language, zhong2022dialoglm}.

\begin{table}[t!]
    \fontsize{7.5}{9} \selectfont
    \centering
    \def\arraystretch{1.2}
    \resizebox{\columnwidth}{!}{
    \begin{tabular}{l}
    \toprule
    \textbf{Meeting Transcript about Fundamental Education} \\
    \toprule
    \tabincell{l}{\textit{Speaker1}:~我想请问一下，我们平时对于基础教育，包括现在因为就\\是只有哈工大有本科，所以\textcolor{mygreen}{哈工大是直接有对接这个本科高中生的}\\\textcolor{mygreen}{这个需求的}，在这方面有没有一些围绕着这个方面的分享。\\
    \textit{Speaker2}:~\textcolor{mygreen}{深圳这个教育和医疗是两个短板}，尤其教育的话这个尤其\\是基础教育，现在高端教育慢慢这个步伐已经开始在加快了。但是\\在基础教育这块~...\\
    \textit{Speaker3}:~另一方面\textcolor{mygreen}{跟家长的这种这个观念也比较有关系}。因为~...\\\textcolor{myblue}{\texttt{[EOS]}}} \\
    \hdashline
    \tabincell{l}{
    \textbf{Headline}:~基础教育的短板\\
    \textbf{Segmentation summary}:~目前基础教育是短板，处于未解决温饱问\\题阶段，学校重视不够。家长的观念也有影响，一直都在走应试教\\育的老路，最终目标都是在高考中去考高分~...\\
    } \\
    \midrule
    \tabincell{l}{\textit{Speaker1}:~老师们讲得特别全面系统，而且把我们整个工作的衔接都\\联系起来了。接下来请~...\\
    \textit{Speaker4}:~我说一下我对人才培养的这个看法~...~\textcolor{myblue}{\texttt{[EOS]}}\\
    }\\
    \hdashline
    \tabincell{l}{
    \textbf{Headline}:~人才培养的方式\\
    \textbf{Segmentation summary}:~要提供一个平台，为青少年的特长提供机\\会，同时还能培养特长，要从小发掘小朋友的特长,~...~，同时新技\\术新思想也要通过大城市蔓延到小城市。
    }\\
    \midrule
    \textbf{...} \\
    \midrule
    \tabincell{l}{\textbf{Overall summary}: ~基础教育要多提供平台，重视基础教育的同时，\\也要多关注孩子的兴趣，学校要提供层次的选人标准，不要太局限\\在成绩上，也要有差异化~...}\\
    \bottomrule
    \end{tabular}}
    \caption{An example from our dataset. The green texts are the highlighted sentences. The token \textcolor{myblue}{\texttt{[EOS]}} is used to distinguish different segmentations. We provide multi-granularity summaries to a meeting transcript, including headline, segmentation summary and overall summary. See the English example in Appendix \ref{sec:trasn_exa}.}
    \label{tab:illustration}
\end{table}

Compared to standard text summarization \citep{nallapati2016abstractive, narayan2018don}, meeting summarization is a much more challenging task because it has more informal and oral expressions, topic shifts, multiple participants and longer context. For this reason, existing datasets for meeting summarization, i.e., AMI \citep{mccowan2005ami} and ICSI \citep{janin2003icsi}, can hardly be used to train a robust summarization model owing to 1) their small size -- AMI and ICSI only contain 137 and 59 pairs of meeting transcripts and summaries, respectively; 2) the specific domain -- AMI just focuses on the product design and development while ICSI concentrates on the academic discussions; 3) coarse-grained summaries -- the summaries in these two datasets are directly written for the whole meeting that might involve various topics.
Although some larger dialogue summarization datasets have been created, e.g., SAMSum \citep{gliwa2019samsum}, DialogSUM \citep{chen2021dialogsum} and MediaSum \citep{zhu2021mediasum}, their context and summaries are still much shorter than the meeting summarization data.
On the other hand, some variants of the summarization task are recently studied in news summarization, such as segmentation-based summarization \citep{liu2022end} aiming at jointly segmenting and summarizing the lengthy context, multi-granularity summarization \citep{zhong2022unsupervised} aiming at generating coarse-, middle- and fined-grained summaries at the same time, and \textit{retrieval-then-generate} summarization \citep{mao2022dyle} aiming at generating the summaries on the extracted salient sentences.
While these variant tasks and methods have demonstrated their capacities on improving the news summarization performance, there is a huge demand for a large-scale \textit{versatile} meeting summarization dataset to adapt these variants into the field.

To this end, we collect a \textbf{V}ersatile \textbf{C}hinese meeting \textbf{Sum}marization dataset based on the real-life meeting recordings, called \textsc{VCSum}. The dataset contains 239 meetings, with a total duration of over 230 hours, and each meeting transcript has over 14K tokens on average. To make the dataset versatile, we provide various annotations, including segmentation-based annotations, multi-granularity annotations and extractive annotations. Table \ref{tab:illustration} illustrates an example from our dataset. The meeting transcript is segmented into several sections according to the topics discussed. Then multi-granularity summaries are provided for each section, i.e., a coarse-grained headline summary with 5-20 words and a find-grained segmentation summary with 100-150 words. An overall summary with 200-250 words for the whole meeting is also annotated to formulate the challenging summarization tasks, such as lengthy meeting transcript summarization.
Furthermore, we instruct annotators to highlight salient sentences in the meeting whose content is further verified to be highly consistent with the summaries. As such, our dataset is also a good testbed for the \textit{extract-then-generate} summarization. In the experiment part, we evaluate several benchmark models on segmentation-based summarization, multi-granularity summarization, extract-then-generate summarization and highlight sentence extraction. We also provide a conversation solution to the dataset.

We summarize our contributions as follows: (1) we are the first one to collect a large and high-quality meeting summarization dataset from real-life videos in the last around 20 years; (2) we propose the first \textit{versatile} summarization dataset that contains the annotations of extractive highlight sentences, topic segmentation, and multi-granularity summaries; (3) we conduct extensive experiments on the proposed dataset, constructing the benchmark to facilitate further research.

\section{Related Work}
\subsection{Dialogue Summarization} Dialogue summarization aims to extract or summarize the most important information in the dialogue. Dialogue can take many forms, including chit-chat, meetings, and emails \citep{li2017dailydialog, zhang2021emailsum, chen2021dialogsum}. A bunch of chat summarization datasets \citep{gliwa2019samsum, zhu2021mediasum, chen2021dialogsum} have been created yet. For example, the most widely used chat summarization dataset, i.e., SAMSum \citep{gliwa2019samsum}, is collected by linguists writing messenger-like conversations. This high-quality dataset greatly facilitates this research direction. A set of follow-up algorithms are proposed to solve the task by enhancing the dialogue context modeling\citep{chen2020multi, zhong2022dialoglm} or dialogue participant modeling \citep{narayan2021planning}. However, due to the much higher cost of collecting and annotating meetings, existing meeting summarization datasets are really limited, only AMI \citep{mccowan2005ami} and ICSI \citep{janin2003icsi}, which were constructed around 20 years ago. The small size and inferior annotation quality of these two datasets are scarce to support the training of a robust meeting summarization model, while meeting summarization suffers more challenges than chat summarization, e.g., longer context and more topic shifts. Although some data augmentation techniques\citep{zhu2020hierarchical,zhong2022dialoglm} have been attempted, insufficient meeting summarization data remains a great barrier to this field moving forward. In this work, we create a high-quality and larger meeting summarization dataset to fill the gap of no new meeting summarization datasets proposed in the last 20 years.

\subsection{Advanced Summarization Tasks}
Thanks to the abundant data and simpler annotations, many variants of the standard summarization task have been attempted on news summarization, including segmentation-based summarization \citep{liu2022end}, multi-granularity summarization \citep{zhong2022unsupervised}. The task of segmentation-based summarization is proposed to address the problem of multiple topics discussed in the lengthy document. The multi-granularity summarization aims to provide summaries with different degrees of semantic coverage. Although these tasks could improve the summarization performance, they cannot be tested on meeting summarization due to a lack of data. In this work, we provide the meeting summarization data with segmentation-based and multi-granularity annotations and build the benchmarks of these tasks on meeting summarization. Besides, superior to \textsc{SegNews} dataset \citep{liu2022end}, \textsc{VCSum} focuses on real-life meetings and provides human-annotated summaries.

%Long documents, such as news articles or meeting transcripts, usually discuss multiple topics, causing the huge difficulties for models to directly generate an accurate summary covering all aspects. To this problem, previous solutions may either generate a summary with the partial context \citep{cheng2016neural} or try to partition the multi-topic articles into segments and then independently summarize them \citet{zhu2021mediasum}. More recently, \citet{liu2022end} create a segmentation-based news summarization dataset to solve this task, namely \textsc{SegNews}, wherein each new article are divided into segments and then summarized on segments.
%However, we argue that our dataset is superior and more challenging one compared to \textsc{SegNews} because 1) our data is collected from real-life meetings, which involves more oral expressions than the written texts; 2) summaries in \textsc{VCSum} are written by annotators for this task while \textsc{SegNews} directly uses the subtitles as summaries; 3) we provide multi-granularity summaries.

\begin{figure}[t!]
    \centering
    \includegraphics[width=0.9\linewidth]{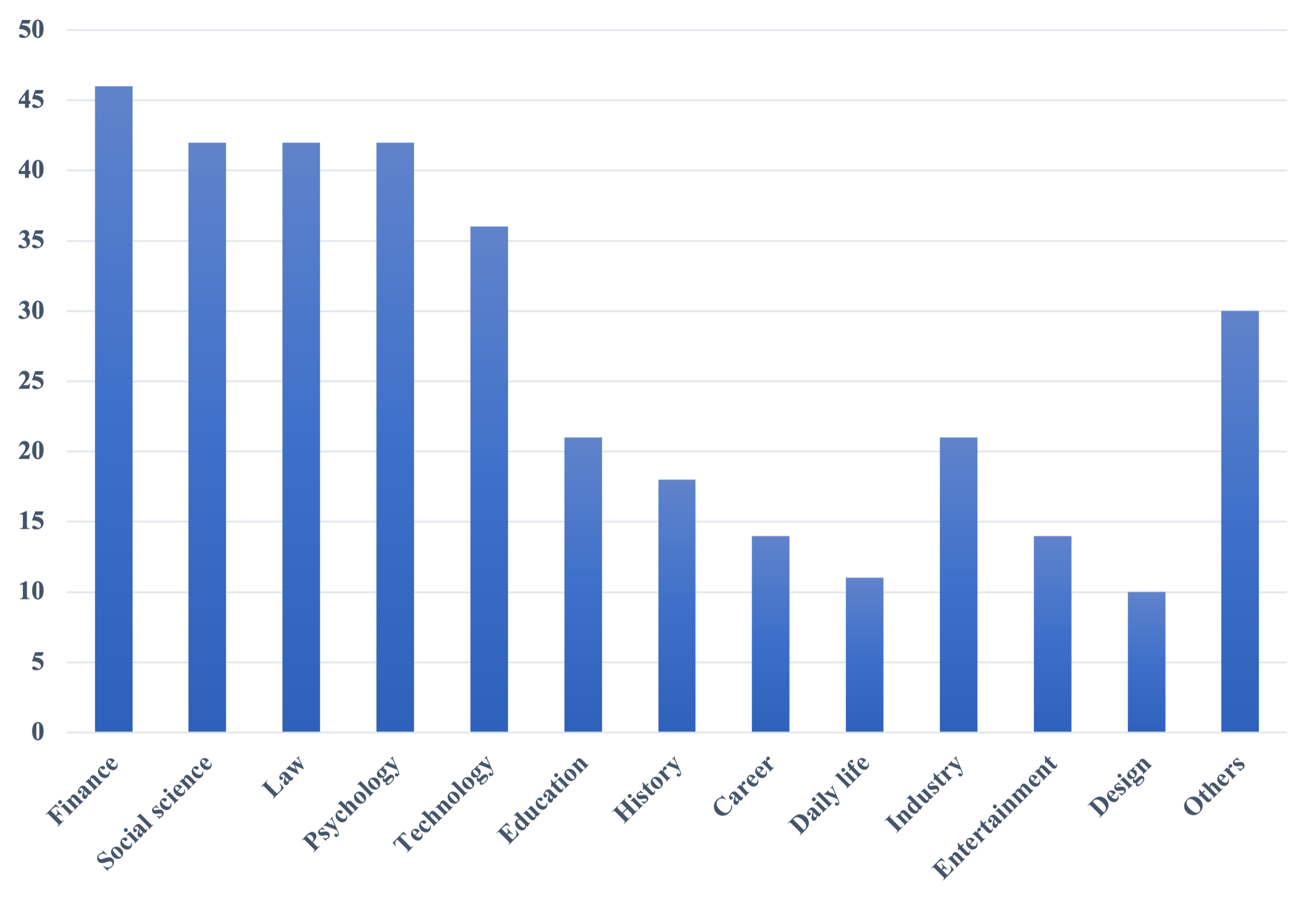}
    \caption{Topic distribution of \textsc{VCSum}.}
    \label{fig:topic_dis}
\end{figure}

\section{The \textsc{VCSum} Corpus}
\subsection{Data Selection} \label{sec:data_selection}
We collect the roundtable meetings from some Chinese video-sharing websites. We first obtain 1,419 videos from the websites by searching the keyword ``圆桌会议 (roundtable)''. To select high-quality videos and alleviate potential ethical issues, we crowdsource the basic meeting information by asking the following questions: 1) are the audio and video of the meeting clear? 2) how many \textit{valid} participants in the meeting? Participants are valid only when they have lots of expressions and clearly articulate their opinions. 3) is there any reference in the meeting? The reference can be the supporting materials of the meeting, such as slides, technical documentation, requirement document, reports, etc. 4) does the meeting involve any offensive or ethical content, including politics, religious issues, gender concerns, or violence?

Each meeting will be marked by two annotators, and the disagreements will be tackled by the third annotator. Then the candidate videos are selected by the conditions of 1) having clear audio and video; 2) the number of valid participants ranging from 1 to 10; 3) no references; 4) not involving any offensive or ethical concerns; and 5) the meeting duration ranging from 10 minutes to 200 minutes. We finally obtained 541 valid videos. The creation year of these videos ranges from 2017 to 2022.
Owing to the limited budget for the annotation, we further selected 239 meetings, covering as many different domains as possible, including technology, finance, daily life and so on. We provide the topic distribution in Figure \ref{fig:topic_dis}.

\subsection{Data Annotation}
The annotation is conducted on Feishu Minutes \footnote{\url{https://meetings.feishu.cn/minutes}}, an integrated platform for video parsing and automatic speech recognition (ASR). We upload candidate videos to the platform and parse them into meeting transcripts. Then annotators are asked to read the transcripts and provide five kinds of annotations.

\paragraph{Highlight sentences.} Annotators should highlight the salient and informative sentences in the original meeting transcripts. The marked sentences must be fluent and complete. To avoid excessive labeling, the highlighted sentences should not exceed 10\% of the entire transcript.

\paragraph{Topic segmentation.} As a meeting generally involves multiple topics, we instruct annotators to identify different topic segments by inserting a special token \textsc{[EOS]} at the end of each segment. Note that the topic segmentation is annotated on the utterance level.

\paragraph{Segmentation headline.} After obtaining topic segments, annotators should provide a headline to identify the topic of each segmentation. The word count for each headline should fall within the range of 5-20 words.

\paragraph{Segmentation summary.} Annotators should write a summary for each segment. Different from headlines, segmentation summaries focus more on details of the content. Each segmentation summary should contain 100-150 words.

\paragraph{Overall summary.} In the end, annotators should provide an overall summary that covers the most salient and informative content of the meeting. The word count for each overall summary should range from 200 to 250 words.

\begin{table*}[t!]
    \fontsize{7.5}{8} \selectfont
    \centering
    % \bgroup
    \def\arraystretch{1.2}
    \begin{tabular}{lcccccccc}
    \toprule
    Dataset & Lan. style & Scenario & Domain & \#Transcripts & \#Tokens/trans. & \#Turns/trans. & \#Speakers/trans. & \#Tokens/sum. \\
    \midrule
    CNN & written & news & multiple & 92465 & 654.0 & - & - & 42.1\\
    DailyMail & written & news & multiple & 219506 & 702.9 & - & - & 51.5\\
    \hdashline
    % CRD3 & written & TV show & game & 159 & 31802.8 & 2507.4 & - & 2062.3\\
    SAMSum & written & online & multiple & 16369 & 93.8 & 11.2 & 2.4 & 20.3\\
    DialogSum & spoken & - & multiple & 13460 & 131.0 & - & - & 23.6\\
    MediaSum & spoken & interview & multiple & 463596 & 1553.7 & 30.0 & 6.5 & 14.4\\
    CSDS & \tabincell{c}{written\\(Chinese)} & online & customer & 10701 & 401.1 & 26.0 & 2.0 & 83.3\\
    \hdashline
    AMI & spoken & meeting & product & 137 & 6007.7 & 535.6 & 4.0 & 296.6 \\
    ICSI & spoken & meeting & academia & 59 & 13317.3 & 819.0 & 6.3 & 488.5 \\
    \hdashline
    \textsc{VCSum} & \multirow{2}{*}{\tabincell{c}{spoken\\(Chinese)}} & \multirow{2}{*}{meeting} & \multirow{2}{*}{multiple} & 239 & 14106.9 & 73.1 & 5.6 & 231.9\\
    \textsc{VCSum}$_{\textit{seg}}$* & & & & 1359 & 2480.9 & 12.9 & 3.0 & 139.1\\
    \bottomrule
    \end{tabular}
    \caption{Comparison between \textsc{VCSum} and other news, dialogue or meeting summarization datasets. * indicates the statistical results on segmentation transcripts. \# stands for the average result.}
    \label{tab:data_stat}
    \vspace{-1em}
\end{table*}

\begin{table}[t!]
    \fontsize{8}{9} \selectfont
    \centering
    \def\arraystretch{1.2}
    \begin{tabular}{lccc}
    \toprule
        \textsc{VCSum} & \#hl./trans. & \#Token/hl. & R-1/2/L \\
        \midrule
        Whole & 71.7 & 32.5 & 88.65/50.46/62.56\\
        Segmentation & 12.7 & 32.5 & 81.17/49.22/67.54\\
    \bottomrule
    \end{tabular}
    \caption{The highlight statistics of the whole meeting transcript and transcript segmentations. hl. stands for highlight sentences. R-1/2/L means the ROUGE-1/2/L recall score between the set of highlights and the corresponding summary.}
    \label{tab:my_label}
\end{table}

\subsection{Quality Control}
To obtain high-quality data, we release a set of data samples to select the most suitable and professional annotators. Finally, we recruit eight annotators majoring in law, finance or Chinese culture from the top universities in China (4 females and 4 males).
Before the formal annotation, all annotators were asked to study the annotation protocols and practice on the training samples for a period of time. The annotation process began after all annotators passed our examination. We also have two annotation inspectors from our research group to monitor the whole process.
During the annotation process, each sample is annotated by an annotator and checked by another annotator and an inspector. The annotation would be accepted only if both two checkers approved it.
After the annotation, all results are further validated by ourselves. If any errors are found in an annotation batch, the corresponding annotator and checkers would be instructed to self-check and re-annotate the batch until the result meets our requirements.
Additionally, to alleviate the error propagation from ASR, we manually compared several ASR technologies and finally selected Feishu Minutes.
Our analysis of 100 randomly sampled meeting segmentations reveals a word error rate of 8\%, which is significantly lower than the error rate of 30\% reported in the AMI dataset. To further improve the data quality, we encourage annotators to revise any typos or grammatical errors they find during the annotation process.

\begin{figure}[t!]
    \centering
    \includegraphics[width=1.0\linewidth]{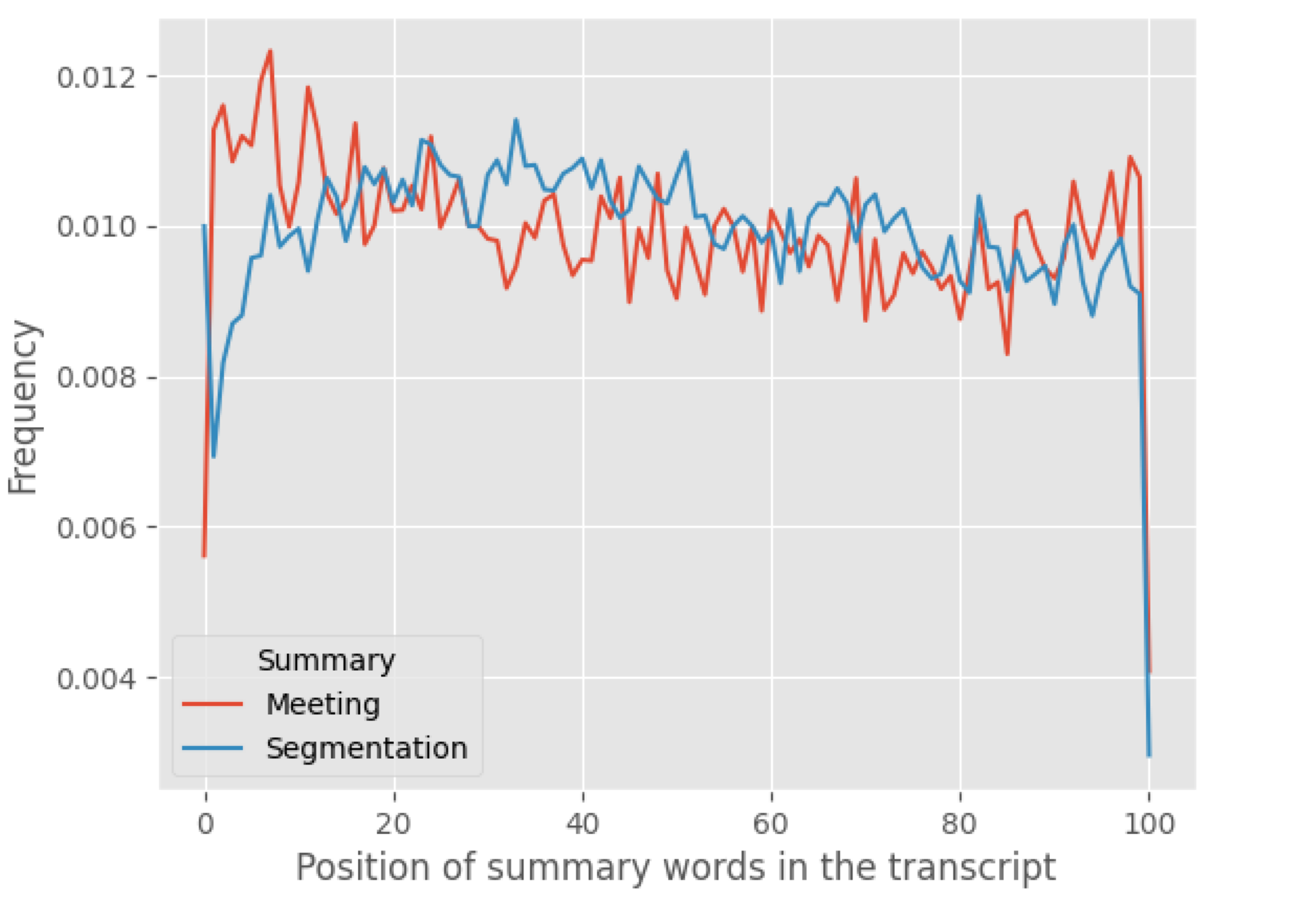}
    \caption{The frequency of non-stop summary words appearing at different positions of the transcript. The positions are normalized to [0, 100].}
    \label{fig:position_bias}
    \vspace{-1.5em}
\end{figure}

\subsection{The Characteristics of \textsc{VCSum}} \label{sec:characteristics}
Table \ref{tab:data_stat} illustrates the comparison between \textsc{VCSum} and other news, dialogue or meeting summarization datasets. We see that \textsc{VCSum} has much longer context and summaries compared to existing news and dialogue summarization datasets. The transcripts in \textsc{VCSum} contain 72.8 turns and 5.7 speakers on average, suggesting that the dataset also has the characteristics of multi-turn and multi-party dialogues. While the traditional meeting datasets, like AMI and ICSI, only focus on a single domain, \textsc{VCSum} involves multiple domains, ranging from daily life to academic discussions.
Moreover, we also provide the extractive highlight sentences in \textsc{VCSum}. Each meeting transcript averagely has 71.7 highlighted sentences, with each sentence containing 32.5 tokens. We calculate the ROUGE scores \citep{lin2004rouge} between highlight sentences and the corresponding summary. The larger value of ROUGE-1 indicates that the written summaries are semantically consistent with the highlight annotations, while the smaller ROUGE-2 scores suggest the abstractiveness of the summary.

Previous study \citep{kedzie2018content} reveals a critical problem of summarization task, called \textit{positional bias}, which is that in existing datasets \citep{mccowan2005ami, chen2021dialogsum,zhu2021mediasum}, most important information is often shown at the beginning of the context. This problem might bias the model to focus on the early words rather than the entire context.
To this end, we also study the positional bias on our dataset. We evenly partition the transcript into 100 bins and count the frequency of the non-stop summary words appearing in each bin. As shown in Figure \ref{fig:position_bias}, the meeting transcripts contain more summary words near the beginning and end while segmentation transcripts hold more summary words in the middle part. However, the summary words of \textsc{VCSum} are smoothly distributed in the transcripts. This observation indicates that our dataset does not suffer the positional bias, thus being a more challenging dataset.

\section{Task Overview}
Based on the annotation of our dataset, we propose three main challenging tasks. Given a meeting transcript $C=\{U_1, U_2, ..., U_N\}$ consisting of $N$ utterances, wherein each utterance $U_i=\{w^i_1, w^i_2, ..., w^i_{|U_i|}\}$ is a sequence of words, we formulate the tasks as follows.

\paragraph{Highlight Sentence Extraction}
The task of highlight sentence extraction (HSE) aims to find a set of spans $H = \{(w^i_j, w^i_k)|0 < i \leq N, 0 < j < k \le |U_i|\}$ that contain the most important information of the meeting. All highlight sentences are not overlapped and not across different utterances.

\paragraph{Segmentation-based Multi-granularity Summarization}
This task is essentially comprised of two sub-tasks, i.e., topic segmentation and multi-granularity summarization. Formally, we aim to generate three summaries at different granularities, i.e., a headline $Y_h$, a segmentation summary $Y_s$ and a joint summary \texttt{$Y_h$:$Y_s$}, based on the transcript segments $S$, where $S \in \{(U_j, ..., U_k)|0 < j \leq k \leq N\}$. The segments are partitioned in the level of utterance, and there are no overlapping among different segments.

\paragraph{Abstractive Meeting Summarization}
The goal of this task is to generate the overall summary $Y_{gold}$ covering the diverse topics in the meeting based on the transcript $C$.

\section{Experiments}
To jointly accommodate the segmentation-based summarization and overall meeting summarization, we divide the train/dev/test datasets that contain around 80\%/10\%/10\% segmentation summaries as well as overall summaries. There are totally 193/25/21 overall summaries and 1076/135/136 segmentation summaries in train/dev/test sets.

\begin{table}[t!]
    \fontsize{10}{11} \selectfont
    \centering
    \def\arraystretch{1.2}
    \begin{tabular}{cccc}
        \toprule[0.8pt]
        Model & Chunk & F1 & Gold R1 \\
        \midrule[0.8pt]
         \multirow{3}{*}{BERT} & 512 & \textbf{39.47} & 84.72 \\ 
         & 1024 & 35.41 & 82.28 \\ 
         \hline
         \multirow{3}{*}{Longformer} & 512 & 39.24 & \textbf{85.17} \\ 
         & 1024 & 38.31 & 84.35 \\ 
         & 2048 & 37.31 & 84.16 \\
        \bottomrule[0.8pt]
    \end{tabular}
    \caption{The F1 and gold ROUGE-1 recall scores on highlight sentence extraction task. BERT and Longformer are evaluated on the data with chunk size of 512, 1024 and 2048.}
    \label{tab:highlight_res}
    \vspace{-1em}
\end{table}

\subsection{Highlight Sentence Extraction}

\paragraph{Experiment Setup} We solve the task as a sequence labeling problem. Due to the lengthy context of meeting transcripts, we try to split the transcript into small chunks. Here we evaluate the BERT \citep{devlin2019bert} and Longformer \citep{Beltagy2020Longformer} models with chunk sizes of 512, 1024 and 2048. We report the F1 score and gold ROUGE-1 recall score as our evaluation metrics, wherein the gold ROUGE recall score is calculated between the generated highlight sentences and the gold overall meeting summary. Find the implementation details in Appendix \ref{sec:impl}.

\paragraph{Results} Table \ref{tab:highlight_res} illustrates the results of highlight sentence extraction on the test set. We can see that both BERT and Longformer perform better with the shorter input length. However, a huge performance drop is observed when BERT works on the longer inputs while Longformer could perform stably. This finding is consistent with the characteristics of these two models.

\begin{table}[t!]
    \fontsize{10}{11} \selectfont
    \centering
    \def\arraystretch{1.2}
    \begin{tabular}{cccc}
        \toprule[0.8pt]
        Model & Chunk Turns & \textit{Pk}$\downarrow$ & \textit{WinDiff}$\downarrow$ \\
        \midrule[0.8pt]
        \textsc{Random} & - & 0.421 & 0.555 \\
        \hline
        \textsc{Even} & - & 0.441 & 0.505 \\
        \hline
         \multirow{4}{*}{\textsc{BertSumExt}} & 5 & 0.293 & 0.310 \\ 
        %  & 8 & 0.240 & 0.242 \\ 
         & 10 & \textbf{0.216} & \textbf{0.214} \\ 
         & 15 & 0.230 & 0.234 \\
         \hline
         \multirow{3}{*}{BART$_{enc}$} & 5 & 0.334 & 0.380 \\ 
         & 10 & 0.222 & 0.216 \\ 
         & 15 & 0.244 & 0.242 \\
        \bottomrule[0.8pt]
    \end{tabular}
    \caption{The \textit{Pk} and \textit{WinDiff} scores on the topic segmentation task. BERT and BART$_{enc}$ are evaluated on the data with chunk turns of 5, 10 and 15, wherein BART$_{enc}$ stands for using only the encoder part of BART model. $\downarrow$ means lower is better.}
    \label{tab:seg_res}
    \vspace{-1em}
\end{table}

\subsection{Segmentation-based Multi-granularity Summarization}

\begin{table*}[t!]
    \fontsize{8.5}{9} \selectfont
    \centering
    \def\arraystretch{1.2}
    \begin{tabular}{cccccccccccccc}
        \toprule[0.8pt]
        \multicolumn{2}{c}{Model} & & \multicolumn{3}{c}{Headline} & & \multicolumn{3}{c}{Segmentation Summary} & & \multicolumn{3}{c}{Joint Summary} \\
        \cline{1-2} \cline{4-6} \cline{8-10} \cline{12-14}
        Segmentor & Generator & & R1 & R2 & RL & & R1 & R2 & RL & & R1 & R2 & RL \\
        \midrule[0.8pt]
        \multicolumn{14}{c}{\textit{With gold segments}} \\
        \hline
        - & \textsc{Random} & & 21.21 & 2.53 & 1.03 & & 43.22 & 7.63 & 13.41 & & 43.18 & 7.66 & 13.49\\
        - & \textsc{Oracle} & & 39.90 & 20.01 & 34.72 & & 55.10 & 25.19 & 32.66 & & 55.21 & 24.82 & 31.77 \\
        % - & {TopicRank} \\
        - & BART($l=1024$) & & 42.85 & 25.53 & 35.96 & & 58.18 & 23.56 & 29.65 & & 59.49 & 24.38 & 29.80 \\
        - & BART($l=2048$) & & 41.92 & 23.73 & 34.53 & & 59.06 & 24.35 & 29.25 & & 59.14 & 24.35 & 29.25 \\
        - & Pegasus($l=1024$) & & \textbf{46.49} & \textbf{27.69} & 38.92 & & 59.31 & 25.10 & 33.19 & & 59.78 & 25.66 & 34.11 \\
        - & Pegasus($l=2048$) & & 45.68 & 27.22 & \textbf{39.04} & & \textbf{59.59} & \textbf{25.31} & \textbf{33.57} & & \textbf{59.80} & \textbf{26.03} & \textbf{34.55} \\
        \midrule[0.8pt]
        \multicolumn{14}{c}{\textit{With predicted segments}} \\
        \hline
        \textsc{BertSumExt} & BART$_{ed}(l=1024)$ & & 40.10 & 21.92 & 32.43 & & 56.63 & 22.17 & 25.69 & & 57.60 & 22.73 & 27.16 \\
        \textsc{BertSumExt} & Pegasus$(l=1024)$ && 41.41 & 22.43 & 34.93 && 57.90 & 23.45 & 30.98 && 58.14 & 23.39 & 30.80 \\
        BART$_{enc}$ & Pegasus$(l=1024)$ && 40.14 & 20.13 & 32.76 & & 54.16 & 21.15 & 26.89 & & 54.01 & 20.97 & 27.51 \\
        BART$_{enc}$ & BART$_{ed}$ & & 37.44 & 19.74 & 30.40 & & 53.30 & 20.12 & 23.46 & & 54.20 & 20.33 & 23.90\\
        \bottomrule[0.8pt]
    \end{tabular}
    \caption{Evaluation results of ROUGE F1 on segmentation-based multi-granularity summarization. The scores here is calculated without sentence splitting. $l$ stands for the truncation length. BART$_{ed}$ means the standard encoder-decoder-based BART model.}
    \label{tab:seg_sum_res}
    \vspace{-1em}
\end{table*}

\paragraph{Experiment Setup}
This task is comprised of two sub-tasks, i.e., topic segmentation and multi-granularity summarization, wherein the summarization method is applied on the segmented sections.

For the topic segmentation, we formulate the task as a sequence labeling problem. We solve it by \textsc{BertSumExt} model \citet{liu2019text}  and BART encoder model \citep{lewis2020bart}. Specifically, we insert a special token \texttt{[CLS]} at the beginning of each utterance, which would be used to classify whether the utterance is the end of a segment. Then we use an interval segmentation indicator to distinguish different utterances. However, due to the context length exceeding 10K in most samples, we split the meeting transcript into chunks where each chunk contains 5/10/15 turns of utterance. In the stage of inference, we make the predictions on segmented chunks but calculate the scores on the entire meeting transcript. Following previous work \citep{zhong2022dialoglm}, we use the standard metrics \textit{Pk} \citep{beeferman1999statistical} and \textit{WinDiff} \citep{pevzner2002critique} to evaluate the segmentation models.

For multi-granularity summarization, we evaluate two settings, i.e., \textit{with predicted segments} and \textit{with gold segments}. For the former setting, the model is trained to first segment the transcript and then generate summaries for the predicted segments.
For the latter, we provide the gold segments and evaluate the capacities of generating multi-granularity summaries with correct segments.
As the number of predicted segments is uncertain, we calculate the ROUGE-1/2/L scores for each meeting by joining all predicted segmentation summaries together with a special token \texttt{[Y\_SEP]}. We employ two widely-used summarization models, i.e., BART and Pegasus \citep{zhang2020pegasus}, as our backbone.
We also report the results of \textsc{Random} and \textsc{Oracle} baselines. \textsc{Random} is to randomly select sentences from the context as the summary, while \textsc{Oracle} is to select the sentences with the highest ROUGE-1 scores against the ground truth.

In the setting of \textit{with predicted segments}, we adopt the best-performing topic segmentation model, i.e., \textsc{BertSumExt} trained with 10-turn chunks, to segment the transcripts. Then the summaries are generated by BART and Pegasus models on the segmented sections. Furthermore, we also try to initialize the encoder part of the generative BART with the weights from BART$_{enc}$. In this way, we expect the generative model can be aware of the segmentation features. Note that we do not follow the previous work \citep{liu2022end} which jointly optimizes the segmentation and generation tasks with the pre-trained language models because our transcripts and summaries are much longer than theirs and the model cannot work well with the such lengthy context.

\paragraph{Results} Table \ref{tab:seg_res} illustrates the evaluation results of topic segmentation. As we can see, both \textsc{BertSumExt} and BART models are much better than the baselines, i.e., \textsc{Random} and \textsc{Even}. \textsc{Random} is the baseline that randomly selects the utterances as the boundary of a segment, while \textsc{Even} is to partition the whole transcript evenly.
\textsc{BertSumExt} outperforms the BART$_{enc}$ across the board. This is reasonable since \textsc{BertSumExt} captures more inter-segment features by the interval indicator and high-level interact layers. The results of these two models also show the same trend that segmenting the transcript into 10 turns is the best choice. This is consistent with the findings in \citet{sankar2019neural} that around 8 turns of dialogue context are enough to capture the contextual features.

Table \ref{tab:seg_sum_res} shows the evaluation results on segmentation-based multi-granularity summarization. When given gold segments, Pegasus $(l=2048)$ achieves the best performance on most metrics. Both BART and Pegasus models outperform the extractive oracle methods across the board, suggesting that the written summaries are abstractive enough. For the summary generation at different granularities, we find that the headline generation is harder than the summary generation owing to its highly condensed information. Comparing the generation of segmentation summary and joint summary, slight improvements are spotted when prompting the segmentation summary generation with the headline.

When evaluating with the predicted segments, we find that comparable performance could be reached with a strong topic segmentation model. After conducting more detailed analyses regarding the results of segmentation and generation, we surprisingly find that the most errors of our segmentation model are within three utterances, e.g., the label is the 10$^{\textit{th}}$ utterance but the prediction is 8$^{\textit{th}}$, while the generation model can easily tolerate such deviations. An exception is tying the weights of the segmentation model with the generation model, i.e., BART$_{enc}$ + BART$_{ed}$, which performs much worse than others. We think this is because the generation process is not largely dependent on the segmentation features.

\begin{table}[t!]
    \fontsize{9}{10} \selectfont
    \centering
    \def\arraystretch{1.2}
    \begin{tabular}{cccc}
    \toprule[0.8pt]
        Method & R1 & R2 & RL \\
    \midrule[0.8pt]
        Vanilla BART$(l=1024)$ & 37.61 & 11.38 & 18.16\\
        Vanilla BART$(l=2048)$ & 38.83 & 12.05 & 18.02\\
        Vanilla Pegasus$(l=1024)$ & 29.43 & 17.19 & 19.42\\
        Vanilla Pegasus$(l=2048)$ & 27.62 & 8.56 & 18.49\\
        \hline
        Pred. Joint Summary + BART & 42.29 & 13.77 & 19.17\\
        Pred. Highlights + BART & \textbf{45.76} & \textbf{16.33} & \textbf{22.26}\\
        \hline
        Gold Joint Summary + BART & 55.85 & 31.33 & 34.19\\
        Gold Highlights + BART & 47.75 & 18.47 & 24.14\\
    \bottomrule[0.8pt]
    \end{tabular}
    \caption{Evaluation results on abstractive meeting summarization. Pred. joint summary/highlights means the pipeline method that first predicts the joint summaries/highlights and then generates the overall summary on the results of first step.}
    \label{tab:abs_res}
    \vspace{-1.5em}
\end{table}

\subsection{Abstractive Meeting Summarization}
\paragraph{Experiment Setup} It is a kind of \textit{long text summarization} task \citep{liu2020clts, gidiotis2020divide}, which encourages the model to generate a concise summary (< 512 words) containing the main content of a long text (> 5,000 words). We evaluate the vanilla sequence-to-sequence models and \textit{retrieval-then-generate} methods. Specifically, we fine-tune the BART and Pegasus models with the truncation length of 1024 and 2048 to generate the overall summary. For \textit{retrieval-then-generate} methods, we first \textit{retrieve} the joint summaries or highlight sentences of a meeting transcript, and then produce the overall summaries on the retrieval results. This method hugely decreases the input length to the summarization model. We use the best-performing highlight model and joint summary generation model to finish the retrieval process. We report the ROUGE-1/2/L F1 scores for performance comparison.

\begin{table*}[t!]
    \fontsize{8.6}{9.5} \selectfont
    \centering
    \begin{tabular}{cc}
    \toprule[0.8pt]
        \multicolumn{2}{c}{Meeting ID: 71962448;~Segment ID: 2} \\
        \hdashline
        \specialrule{0em}{1.5pt}{1.5pt}
        \textbf{Pegasus} & \tabincell{l}{{中国当代艺术蓬勃发展的原因}:中国有庞大的艺术教育的底盘，学校的空间很大，可以让新的年轻艺术家\\有很好的活动的条件，中国的材料很丰富，加工便宜，人力资源很便宜，在市场上还有快速发展的过程，\\至少有潜在的购买力吸引这个艺术家，在世界议坛上很流行的中国政治口号，在文化信息的交往中，中\\国艺术家比西方艺术家有更宽拓的空间。}  \\
        \specialrule{0em}{1.5pt}{1.5pt}
        \textbf{Gold} & \tabincell{l}{艺术教育问题的分析:从艺术教育的角度来看，{中国当代艺术蓬勃发展有六大原因}，一是{中国有庞大的艺}\\{术教育地盘}，二是{学校空间大，可以让新的年轻艺术家有很好的活动条件}，三是{中国的材料丰富，并且}\\{人力资源便宜}，四是{市场上有快速发展的过程，存在潜在的购买力}，五是{中国的元素，文化信息交流中，}\\{中国艺术家更具宽拓空间}，六是\textcolor{myred}{中国正处于巨大变革中，批判性的建构是当代艺术的重要动力。}} \\
        \midrule[0.8pt]
        \multicolumn{2}{c}{Meeting ID: 634267454} \\
        \hdashline
        \specialrule{0em}{1.5pt}{1.5pt}
        \textbf{\tabincell{c}{Pred. Sum.\\+ BART}} & \tabincell{l}{\textcolor{myblue}{区块链是一个学科化的研究而不仅仅是热点}，它是一种从特殊走向另一种特殊的时代，它一直尝试对社\\会进行范式的迁移，从历史交汇到文化主导，在未来艺术形式的一个很关键的线索，可以发现不断不断\\地将权力分散到更多人的手中，能够参与到社会的运动中，这也是未来的艺术选择艺术家的标准。艺术\\作品的表达方式可以用画面表达和用音像的方式表达，或者是用任何一种行为方式，关键是能否用擅长\\的工具、或者愿意拓展你的想象力去表达你对加密世界的思考。}\\
        \specialrule{0em}{1.5pt}{1.5pt}
        \textbf{Gold} & \tabincell{l}{加密的爆发搅活了整个艺术品市场，带动了艺术领域改革开放，打破了原有的概念，带来了新的创作媒\\介，是对艺术行业的一个挑战，是一个转向，让创作者与世界更好的融合，但是艺术家还是要有标准的\\，要有人品、有创新的能力、要让作品更够更清晰的表达，加密是从底层起来的，有着蓬勃的生命力，\\每一个参与社会推进的人就已经是在参与艺术的创作了。}\\
    \bottomrule[0.8pt]
    \end{tabular}
    \caption{Case studies on the joint summary (the first part) of the segment and the overall summary (the second part). Find the translations in Appendix \ref{sec:trans_case}.}
    \label{tab:case_study}
    \vspace{-1em}
\end{table*}

\begin{table}[t!]
    \fontsize{8.5}{9} \selectfont
    \centering
    \def\arraystretch{1.2}
    \begin{tabular}{cccc}
    \toprule[0.8pt]
        Method & Headline & Seg. Sum & Joint Sum \\
    \midrule[0.8pt]
        BART$(l=2048)$ & 41.92 & 59.06 & 59.14\\
        Pegasus$(l=2048)$ & \textbf{45.68} & 59.59 & 59.80\\
        \hline
        \textsc{ConvLM}$(l=2048)$ & 45.25 & \textbf{59.98} & \textbf{60.70} \\
    \bottomrule[0.8pt]
    \end{tabular}
    \caption{ROUGE-1 F1 scores of different models on multi-granularity summarization with gold segments.}
    \label{tab:convlm_res}
    \vspace{-1em}
\end{table}

\paragraph{Results} As shown in Table \ref{tab:abs_res}, the vanilla models perform weakly on meeting summarization owing to the lengthy input. As said in Section \ref{sec:characteristics}, the summary words are evenly distributed in the entire transcript. Therefore, directly truncating the input would cause information loss. The paradigm of \textit{retrieval-then-generate} is a good solution to this problem since it retrieves the key information first and then generates the summary on the retrieval results.
When the summarization is based on gold content, the joint summary is a better ground than the highlight information. This is reasonable since the joint summary contains more concentrated information with fewer words. However, when the summarization is based on the retrieval results, predicted highlights even achieve comparable summarization performance to the gold highlights. This is attributed to the effectiveness of our highlight extraction model and the redundancy information of highlight sentences. The worse performance from the predicted joint summary is because the generation of joint summaries is essentially a challenging task, which might cause more error propagation.

\section{Further Discussions}
\subsection{Conversational Solutions}
\textsc{DialogueLM} \citep{zhong2022dialoglm} is a strong baseline for meeting summarization in English. Based on the fact that meeting is a kind of long conversation, \textsc{DialogueLM} adopts sparse architecture and dialogue-specific pre-training objectives, such as speaker masking, turn splitting, turn merging and turn permutation, to capture the conversational features. It finally demonstrates that the pre-trained dialogue model is also a good solution to the meeting summarization task. Inspired by this observation, we make a preliminary attempt at the conversational solution to \textsc{VCSum}.
Specifically, we pre-train an encoder-decoder-based dialogue model, dubbed \textsc{ConvLM}, using our in-house Chinese dialogue data. We provide details of \textsc{ConvLM} in Appendix \ref{sec:convlm}. We train the model with the objectives of speaker identification and response generation. After the pre-training, we fine-tune the model on \textsc{VCSum}. To simplify the comparison, we evaluate on the task of multi-granularity summarization with gold segments.

As shown in Table \ref{tab:convlm_res}, \textsc{ConvLM} achieves comparable or better performance against the summarization models. Especially for the generation of segmentation and joint summary, \textsc{ConvLM} consistently outperforms the baseline models, suggesting that modeling conversational features could benefit the summarization task on \textsc{VCSum}. This finding demonstrates the conversational characteristics of our dataset and sheds light on using dialogue-specific pre-trained language models to solve the tasks of \textsc{VCSum}.

\subsection{Error Analysis}
To study the difficulties of \textsc{VCSum}, we take a detailed analysis of the error cases. We find that around 50\% of errors is information missing which is essentially caused by lengthy input truncation. For example, in the first row of Table \ref{tab:case_study}, the segmentation summary is much semantically close to the ground truth with only one key point (the red words) missing. This phenomenon is more severe when generating the overall summaries owing to the long and informative context. There are also some errors from irrelevant information or redundant information, accounting for around 20\%. These kinds of errors are mostly found in the retrieval-then-generate methods that retrieve some insignificant content in the first stage, thus finally misleading the generation process, like the blue words in Table \ref{tab:case_study}. The remaining 30\% errors include factual errors and syntactic errors.

\section{Conclusion and Future Work}
In this work, we collect a large and high-quality Chinese meeting summarization dataset from real-life videos, namely \textsc{VCSum}. The dataset is versatile to support the tasks of highlight sentence extraction, segmentation-based summarization, multi-granularity summarization and meeting summarization. Depth analyses demonstrate the superiority of our dataset. We then provide a strong benchmark for different downstream tasks on \textsc{VCSum}.
For future work, we believe the development of an end-to-end framework that could jointly solve all tasks of \textsc{VCSum} is a promising direction. Furthermore, we will also consider to release the video and audio files of the annotated meetings to facilitate the research of multi-modality meeting summarization.

\section*{Acknowledgement}
We sincerely appreciate the valuable and constructive comments from the reviewers.
This work was supported in part by the Hong Kong ITF grant PRP/079/22FX, the Technological Breakthrough Project of Science, Technology and Innovation Commission of Shenzhen Municipality under Grants JSGG20201102162000001, InnoHK initiative, the Government of the HKSAR, Laboratory for AI-Powered Financial Technologies.

\section*{Limitations}
A potential limitation of this work is that we just try some straightforward methods on the summarization tasks, such as vanilla generative pre-trained language models or pipeline retrieval-then-generate methods. We do not try the end-to-end two-stage approaches in this work since we are more focused on the contributions of the dataset construction and building the benchmark. We leave the development of advanced models as future work.

\section*{Ethics Statement}

\paragraph{The construction of dataset.}
All videos in our newly-introduced dataset are available on the Chinese video sharing websites and are public to the download. To avoid the potential ethical issues, we carefully checked all videos in multiple aspects, as said in Section \ref{sec:data_selection}. We try to guarantee that all videos do not involve any offensive, gender-biased, political content and any other ethical issues. During the annotation, we instruct annotators to anonymize or remove the sensitive or private information.

We recruit eight annotators that passed our examinations from the crowdsourcing platform, and two quality inspectors from our research team. To fairly paid the annotations, we first take an in-house annotation to evaluate the speed and difficulty of the large-scale annotations. Finally, we pay each annotator \$25-\$30 per hour. Typically, it would take around 2 hours to annotate a one-hour meeting. So, the workers are compensated \$50-\$60 per sample.

\paragraph{Applications.}
We will release the code and dataset along with friendly instructions to support its correct use. However, we still need to emphasize that abstractive summarization is a kind of generation task, which is not as controllable as we think. It still would generate some novel or unexpected words occasionally. Therefore, further research on the summarization faithfulness is warmly needed.

% Entries for the entire Anthology, followed by custom entries
\bibliography{custom}

\begin{thebibliography}{31}
\expandafter\ifx\csname natexlab\endcsname\relax\def\natexlab#1{#1}\fi

\bibitem[{Beeferman et~al.(1999)Beeferman, Berger, and
  Lafferty}]{beeferman1999statistical}
Doug Beeferman, Adam Berger, and John Lafferty. 1999.
\newblock Statistical models for text segmentation.
\newblock \emph{Machine learning}, 34(1):177--210.

\bibitem[{Beltagy et~al.(2020)Beltagy, Peters, and
  Cohan}]{Beltagy2020Longformer}
Iz~Beltagy, Matthew~E. Peters, and Arman Cohan. 2020.
\newblock Longformer: The long-document transformer.
\newblock \emph{arXiv:2004.05150}.

\bibitem[{Chen and Yang(2020)}]{chen2020multi}
Jiaao Chen and Diyi Yang. 2020.
\newblock Multi-view sequence-to-sequence models with conversational structure
  for abstractive dialogue summarization.
\newblock In \emph{Proceedings of the 2020 Conference on Empirical Methods in
  Natural Language Processing (EMNLP)}, pages 4106--4118.

\bibitem[{Chen et~al.(2021)Chen, Liu, Chen, and Zhang}]{chen2021dialogsum}
Yulong Chen, Yang Liu, Liang Chen, and Yue Zhang. 2021.
\newblock Dialogsum: A real-life scenario dialogue summarization dataset.
\newblock In \emph{Findings of the Association for Computational Linguistics:
  ACL-IJCNLP 2021}, pages 5062--5074.

\bibitem[{Devlin et~al.(2019)Devlin, Chang, Lee, and
  Toutanova}]{devlin2019bert}
Jacob Devlin, Ming-Wei Chang, Kenton Lee, and Kristina Toutanova. 2019.
\newblock Bert: Pre-training of deep bidirectional transformers for language
  understanding.
\newblock In \emph{Proceedings of the 2019 Conference of the North American
  Chapter of the Association for Computational Linguistics: Human Language
  Technologies, Volume 1 (Long and Short Papers)}, pages 4171--4186.

\bibitem[{Feng et~al.(2021)Feng, Feng, Qin, Qin, and Liu}]{feng2021language}
Xiachong Feng, Xiaocheng Feng, Libo Qin, Bing Qin, and Ting Liu. 2021.
\newblock Language model as an annotator: Exploring dialogpt for dialogue
  summarization.
\newblock In \emph{Proceedings of the 59th Annual Meeting of the Association
  for Computational Linguistics and the 11th International Joint Conference on
  Natural Language Processing (Volume 1: Long Papers)}, pages 1479--1491.

\bibitem[{Gidiotis and Tsoumakas(2020)}]{gidiotis2020divide}
Alexios Gidiotis and Grigorios Tsoumakas. 2020.
\newblock A divide-and-conquer approach to the summarization of long documents.
\newblock \emph{IEEE/ACM Transactions on Audio, Speech, and Language
  Processing}, 28:3029--3040.

\bibitem[{Gliwa et~al.(2019)Gliwa, Mochol, Biesek, and Wawer}]{gliwa2019samsum}
Bogdan Gliwa, Iwona Mochol, Maciej Biesek, and Aleksander Wawer. 2019.
\newblock Samsum corpus: A human-annotated dialogue dataset for abstractive
  summarization.
\newblock In \emph{Proceedings of the 2nd Workshop on New Frontiers in
  Summarization}, pages 70--79.

\bibitem[{Janin et~al.(2003)Janin, Baron, Edwards, Ellis, Gelbart, Morgan,
  Peskin, Pfau, Shriberg, Stolcke et~al.}]{janin2003icsi}
Adam Janin, Don Baron, Jane Edwards, Dan Ellis, David Gelbart, Nelson Morgan,
  Barbara Peskin, Thilo Pfau, Elizabeth Shriberg, Andreas Stolcke, et~al. 2003.
\newblock The icsi meeting corpus.
\newblock In \emph{2003 IEEE International Conference on Acoustics, Speech, and
  Signal Processing, 2003. Proceedings.(ICASSP'03)}, volume~1, pages I--I.
  IEEE.

\bibitem[{Kedzie et~al.(2018)Kedzie, Mckeown, and
  Daum{\'e}~III}]{kedzie2018content}
Chris Kedzie, Kathleen Mckeown, and Hal Daum{\'e}~III. 2018.
\newblock Content selection in deep learning models of summarization.
\newblock In \emph{Proceedings of the 2018 Conference on Empirical Methods in
  Natural Language Processing}, pages 1818--1828.

\bibitem[{Kingma and Ba(2015)}]{kingma2015adam}
Diederik~P Kingma and Jimmy Ba. 2015.
\newblock Adam: A method for stochastic optimization.
\newblock In \emph{ICLR (Poster)}.

\bibitem[{Lewis et~al.(2020)Lewis, Liu, Goyal, Ghazvininejad, Mohamed, Levy,
  Stoyanov, and Zettlemoyer}]{lewis2020bart}
Mike Lewis, Yinhan Liu, Naman Goyal, Marjan Ghazvininejad, Abdelrahman Mohamed,
  Omer Levy, Veselin Stoyanov, and Luke Zettlemoyer. 2020.
\newblock Bart: Denoising sequence-to-sequence pre-training for natural
  language generation, translation, and comprehension.
\newblock In \emph{Proceedings of the 58th Annual Meeting of the Association
  for Computational Linguistics}, pages 7871--7880.

\bibitem[{Li et~al.(2017)Li, Su, Shen, Li, Cao, and Niu}]{li2017dailydialog}
Yanran Li, Hui Su, Xiaoyu Shen, Wenjie Li, Ziqiang Cao, and Shuzi Niu. 2017.
\newblock Dailydialog: A manually labelled multi-turn dialogue dataset.
\newblock In \emph{Proceedings of the Eighth International Joint Conference on
  Natural Language Processing (Volume 1: Long Papers)}, pages 986--995.

\bibitem[{Lin(2004)}]{lin2004rouge}
Chin-Yew Lin. 2004.
\newblock Rouge: A package for automatic evaluation of summaries.
\newblock In \emph{Text summarization branches out}, pages 74--81.

\bibitem[{Liu et~al.(2020)Liu, Zhang, Chen, Cao, and Li}]{liu2020clts}
Xiaojun Liu, Chuang Zhang, Xiaojun Chen, Yanan Cao, and Jinpeng Li. 2020.
\newblock Clts: A new chinese long text summarization dataset.
\newblock In \emph{Natural Language Processing and Chinese Computing: 9th CCF
  International Conference, NLPCC 2020, Zhengzhou, China, October 14--18, 2020,
  Proceedings, Part I}, pages 531--542.

\bibitem[{Liu and Lapata(2019)}]{liu2019text}
Yang Liu and Mirella Lapata. 2019.
\newblock Text summarization with pretrained encoders.
\newblock In \emph{Proceedings of the 2019 Conference on Empirical Methods in
  Natural Language Processing and the 9th International Joint Conference on
  Natural Language Processing (EMNLP-IJCNLP)}, pages 3730--3740.

\bibitem[{Liu et~al.(2022)Liu, Zhu, and Zeng}]{liu2022end}
Yang Liu, Chenguang Zhu, and Michael Zeng. 2022.
\newblock End-to-end segmentation-based news summarization.
\newblock In \emph{Findings of the Association for Computational Linguistics:
  ACL 2022}, pages 544--554.

\bibitem[{Mao et~al.(2022)Mao, Wu, Ni, Zhang, Zhang, Yu, Deb, Zhu, Awadallah,
  and Radev}]{mao2022dyle}
Ziming Mao, Chen~Henry Wu, Ansong Ni, Yusen Zhang, Rui Zhang, Tao Yu,
  Budhaditya Deb, Chenguang Zhu, Ahmed Awadallah, and Dragomir Radev. 2022.
\newblock Dyle: Dynamic latent extraction for abstractive long-input
  summarization.
\newblock In \emph{Proceedings of the 60th Annual Meeting of the Association
  for Computational Linguistics (Volume 1: Long Papers)}, pages 1687--1698.

\bibitem[{McCowan et~al.(2005)McCowan, Carletta, Kraaij, Ashby, Bourban, Flynn,
  Guillemot, Hain, Kadlec, Karaiskos et~al.}]{mccowan2005ami}
I~McCowan, J~Carletta, W~Kraaij, S~Ashby, S~Bourban, M~Flynn, M~Guillemot,
  T~Hain, J~Kadlec, V~Karaiskos, et~al. 2005.
\newblock The ami meeting corpus.
\newblock In \emph{Proceedings of the 5th International Conference on Methods
  and Techniques in Behavioral Research.}, pages 88--100.

\bibitem[{Nallapati et~al.(2016)Nallapati, Zhou, Gulcehre, and
  Xiang}]{nallapati2016abstractive}
Ramesh Nallapati, Bowen Zhou, Caglar Gulcehre, and Bing Xiang. 2016.
\newblock Abstractive text summarization using sequence-to-sequence rnns and
  beyond.
\newblock In \emph{Proceedings of The 20th SIGNLL Conference on Computational
  Natural Language Learning}, pages 280--290.

\bibitem[{Narayan et~al.(2018)Narayan, Cohen, and Lapata}]{narayan2018don}
Shashi Narayan, Shay~B Cohen, and Mirella Lapata. 2018.
\newblock Don’t give me the details, just the summary! topic-aware
  convolutional neural networks for extreme summarization.
\newblock In \emph{Proceedings of the 2018 Conference on Empirical Methods in
  Natural Language Processing}, pages 1797--1807.

\bibitem[{Narayan et~al.(2021)Narayan, Zhao, Maynez, Sim{\~o}es, Nikolaev, and
  McDonald}]{narayan2021planning}
Shashi Narayan, Yao Zhao, Joshua Maynez, Gon{\c{c}}alo Sim{\~o}es, Vitaly
  Nikolaev, and Ryan McDonald. 2021.
\newblock Planning with learned entity prompts for abstractive summarization.
\newblock \emph{Transactions of the Association for Computational Linguistics},
  9:1475--1492.

\bibitem[{Pevzner and Hearst(2002)}]{pevzner2002critique}
Lev Pevzner and Marti~A Hearst. 2002.
\newblock A critique and improvement of an evaluation metric for text
  segmentation.
\newblock \emph{Computational Linguistics}, 28(1):19--36.

\bibitem[{Sankar et~al.(2019)Sankar, Subramanian, Pal, Chandar, and
  Bengio}]{sankar2019neural}
Chinnadhurai Sankar, Sandeep Subramanian, Christopher Pal, Sarath Chandar, and
  Yoshua Bengio. 2019.
\newblock Do neural dialog systems use the conversation history effectively? an
  empirical study.
\newblock In \emph{Proceedings of the 57th Annual Meeting of the Association
  for Computational Linguistics}, pages 32--37.

\bibitem[{Zhang et~al.(2020)Zhang, Zhao, Saleh, and Liu}]{zhang2020pegasus}
Jingqing Zhang, Yao Zhao, Mohammad Saleh, and Peter Liu. 2020.
\newblock Pegasus: Pre-training with extracted gap-sentences for abstractive
  summarization.
\newblock In \emph{International Conference on Machine Learning}, pages
  11328--11339. PMLR.

\bibitem[{Zhang et~al.(2021)Zhang, Celikyilmaz, Gao, and
  Bansal}]{zhang2021emailsum}
Shiyue Zhang, Asli Celikyilmaz, Jianfeng Gao, and Mohit Bansal. 2021.
\newblock Emailsum: Abstractive email thread summarization.
\newblock In \emph{Proceedings of the 59th Annual Meeting of the Association
  for Computational Linguistics and the 11th International Joint Conference on
  Natural Language Processing (Volume 1: Long Papers)}, pages 6895--6909.

\bibitem[{Zhong et~al.(2022{\natexlab{a}})Zhong, Liu, Ge, Mao, Jiao, Zhang, Xu,
  Zhu, Zeng, and Han}]{zhong2022unsupervised}
Ming Zhong, Yang Liu, Suyu Ge, Yuning Mao, Yizhu Jiao, Xingxing Zhang, Yichong
  Xu, Chenguang Zhu, Michael Zeng, and Jiawei Han. 2022{\natexlab{a}}.
\newblock Unsupervised summarization with customized granularities.
\newblock \emph{arXiv preprint arXiv:2201.12502}.

\bibitem[{Zhong et~al.(2022{\natexlab{b}})Zhong, Liu, Xu, Zhu, and
  Zeng}]{zhong2022dialoglm}
Ming Zhong, Yang Liu, Yichong Xu, Chenguang Zhu, and Michael Zeng.
  2022{\natexlab{b}}.
\newblock Dialoglm: Pre-trained model for long dialogue understanding and
  summarization.
\newblock In \emph{Proceedings of the AAAI Conference on Artificial
  Intelligence}, volume~36, pages 11765--11773.

\bibitem[{Zhong et~al.(2021)Zhong, Yin, Yu, Zaidi, Mutuma, Jha, Hassan,
  Celikyilmaz, Liu, Qiu et~al.}]{zhong2021qmsum}
Ming Zhong, Da~Yin, Tao Yu, Ahmad Zaidi, Mutethia Mutuma, Rahul Jha, Ahmed
  Hassan, Asli Celikyilmaz, Yang Liu, Xipeng Qiu, et~al. 2021.
\newblock Qmsum: A new benchmark for query-based multi-domain meeting
  summarization.
\newblock In \emph{Proceedings of the 2021 Conference of the North American
  Chapter of the Association for Computational Linguistics: Human Language
  Technologies}, pages 5905--5921.

\bibitem[{Zhu et~al.(2021)Zhu, Liu, Mei, and Zeng}]{zhu2021mediasum}
Chenguang Zhu, Yang Liu, Jie Mei, and Michael Zeng. 2021.
\newblock Mediasum: A large-scale media interview dataset for dialogue
  summarization.
\newblock In \emph{Proceedings of the 2021 Conference of the North American
  Chapter of the Association for Computational Linguistics: Human Language
  Technologies}, pages 5927--5934.

\bibitem[{Zhu et~al.(2020)Zhu, Xu, Zeng, and Huang}]{zhu2020hierarchical}
Chenguang Zhu, Ruochen Xu, Michael Zeng, and Xuedong Huang. 2020.
\newblock A hierarchical network for abstractive meeting summarization with
  cross-domain pretraining.
\newblock In \emph{Findings of the Association for Computational Linguistics:
  EMNLP 2020}, pages 194--203.

\end{thebibliography}
\bibliographystyle{acl_natbib}

\appendix

\section{Translations} \label{sec:appendix}

\subsection{Translated Examples.} \label{sec:trasn_exa}
See Table \ref{tab:en_illustration}.

\begin{table}[t!]
    \fontsize{8}{9} \selectfont
    \centering
    \def\arraystretch{1.2}
    \resizebox{\columnwidth}{!}{
    \begin{tabular}{l}
    \toprule
    \textbf{Meeting Transcript about Fundamental Education} \\
    \toprule
    \tabincell{l}{\textit{Speaker1}:~I would like to ask about the basic education. Now only Harbin\\ Institute of Technology has undergraduates in Shenzhen. \textcolor{mygreen}{So Harbin Institute}\\ \textcolor{mygreen}{of Technology needs to educate the fresh undergraduates who just finish the}\\ \textcolor{mygreen}{high school courses}. Do you have any sharing regarding this aspect.\\
    \textit{Speaker2}:~\textcolor{mygreen}{Education and medical care are two shortcomings of Shenzen}. E-\\specially in terms of basic education, although the development of high-end \\education has been greatly improved, the basic education ...\\
    \textit{Speaker3}:~\textcolor{mygreen}{On the other hand, it is also related to the thoughts of parents}. Be-\\cause ... \textcolor{myblue}{\texttt{[EOS]}}} \\
    \hdashline
    \tabincell{l}{
    \textbf{Headline}:~Shortcomings of Fundamental Education\\
    \textbf{Segmentation summary}:~At present, basic education is weak, and it is still\\ in the stage of unresolved food and clothing problems, and schools do not\\ pay enough attention to it. The thought of parents also has an influence. Th-\\ey have been following the old path of exam-oriented education, and the ul-\\timate goal is to get high scores in the college entrance examination.....\\
    } \\
    \midrule
    \tabincell{l}{\textit{Speaker1}:~It is a very comprehensive sharing, covering the all aspects of our\\ work. Next, we invite ...\\
    \textit{Speaker4}:~OK, I would like to share my view of talent training ...~\textcolor{myblue}{\texttt{[EOS]}}\\
    }\\
    \hdashline
    \tabincell{l}{
    \textbf{Headline}:~The methods of talent training\\
    \textbf{Segmentation summary}:~It is necessary to provide a platform to provide op-\\portunities for the young people, and at the same time to cultivate their stren-\\gths. It is necessary to discover the strengths of children from an early age. ... \\At the same time, new technologies and new ideas must be spread to small ci-\\ties through big cities.
    }\\
    \midrule
    \textbf{...} \\
    \midrule
    \tabincell{l}{\textbf{Overall summary}: ~Basic education should provide more platforms. While\\ paying attention to basic education, we should also pay more attention to ch-\\ildren's interests. Schools should provide a level of selection criteria, not too\\ limited in grades, but also differentiated~...}\\
    \bottomrule
    \end{tabular}}
    \caption{An example from our dataset in English. The green texts are the highlighted sentences. The token \textcolor{myblue}{\texttt{[EOS]}} is used to distinguish different segmentations. We provide multi-granularity summaries to a meeting transcript, including headline, segmentation summary and overall summary.}
    \label{tab:en_illustration}
\end{table}

\subsection{Translated Case Study} \label{sec:trans_case}
See Table \ref{tab:en_case_study}.

\section{Implementation} \label{sec:impl}
We provide the details of implementations here. For all experiments, we take five runs with different random seeds and report the average score.

\paragraph{Highlight Sentence Extraction} We solve the task at sentence-level. Specifically, we split the utterances into small sentences by \textit{comma}. Then, we insert a special token \textsc{[CLS]} at the beginning of each sentence to represent the sentence. The binary classification is conducted on the special tokens to decide whether a sentence should be highlighted or not. Owing to the unbalanced ratio between positive and negative samples, we only active 40\% negative samples in the training stage. The batch size is set to 16 or 32, depending on the chunk size. The models are optimized by Adam \citep{kingma2015adam} optimizer with an initial learning rate of 5e-5. The pre-trained BERT (\texttt{hfl/chinese-roberta-wwm-ext-large}) and Longformer (\texttt{IDEA-CCNL/Erlangshen- Longformer-330M}) models are loaded from Huggingface\footnote{\url{https://huggingface.co/}}.
We train the models for 10 epochs and select the best model on the validation set to evaluate on the test set. All experiments are conducted on 8 NVIDIA A100 GPUs.

\paragraph{Segmentation-based Multi-granularity Summarization}
For topic segmentation, the \textsc{BertSumExt} is based on the large BERT model while BART$_{enc}$ is based on large BART. We truncate each utterance with maximum utterance length of 128 words. The batch size is set to 16 or 32, depending on the chunk turn size. The other settings are mostly same to highlight sentence extraction.

For multi-granularity summarization, we employ the BART (fnlp/bart-large-chinese) and Pegasus (IDEA-CCNL/Randeng-Pegasus-523M-Summary-Chinese) models from Huggingface as our backbones. The input sequence is truncated to the maximum length of 1024 or 2048. To incorporate speaker information, we add a speaker indicator at the beginning of each utterance, e.g., \textit{[Speaker\_1: utterance\_1; ...;Speaker\_n: utterance\_n;]}. The batch size is set to 64. All models are optimized by Adam, and the learning rate is initialized to 5e-5 and linearly updated. During the training process, the best model is selected on the validation loss. In the stage of inference, we generate the summary using beam search with beam size of 5 and length penalty of 1.0.

\paragraph{Abstractive Meeting Summarization}
For the vanilla generative models, we truncate the input sequence to the maximum length of 1024 or 2048. The batch size is set to 64 or 32, depending on the truncation length. For the retrieval-then-generate methods, all predictions are obtained from the corresponding best-performing models, i.e., \textsc{BertSumExt}+Pegasus$(l=1024)$ and BERT trained with chunk size of 512. During the decoding, we generate the summary using beam search with beam size of 5 and length penalty of 1.2.

\begin{table*}[t!]
    \fontsize{8.6}{9.5} \selectfont
    \centering
    \begin{tabular}{cc}
    \toprule[0.8pt]
        \multicolumn{2}{c}{Meeting ID: 71962448;~Segment ID: 2} \\
        \hdashline
        \specialrule{0em}{1.5pt}{1.5pt}
        \textbf{Pegasus} & \tabincell{l}{Reasons for the vigorous development of Chinese contemporary art: China has a huge infrastructure for art\\ education, and the school has a lot of space, which allows new young artists to have good conditions for act-\\ivities. China's materials are abundant, processing is cheap, human resources are cheap, and there is still ra-\\pid development in the market. At least there is the potential purchasing power to attract the artist. Chinese\\ political slogans are very popular in the world forum. In the exchange of cultural information, Chinese arti-\\sts have a wider space than Western artists.}  \\
        \specialrule{0em}{1.5pt}{1.5pt}
        \textbf{Gold} & \tabincell{l}{Analysis of Art Education Problems:From the perspective of art education, there are six reasons for the vig-\\orous development of Chinese contemporary art. One is that China has a huge territory for art education. Se-\\cond, the school has a large space, which allows new young artists to have good conditions for activities. Th-\\ird, China is rich in materials and cheap in human resources. Fourth, there is a process of rapid development\\ in the market and potential purchasing power. The fifth is Chinese elements. In the exchange of cultural info-\\rmation, Chinese artists have more room to expand. \textcolor{myred}{Sixth, China is undergoing tremendous changes, and cri-}\\ \textcolor{myred}{tical construction is an important driving force for contemporary art}.} \\
        \midrule[0.8pt]
        \multicolumn{2}{c}{Meeting ID: 634267454} \\
        \hdashline
        \specialrule{0em}{1.5pt}{1.5pt}
        \textbf{\tabincell{c}{Pred. Sum.\\+ BART}} & \tabincell{l}{\textcolor{myblue}{Blockchain is a disciplined research rather than just a hotspot}. It is an era from a special to another special,\\ and it has been trying to transfer the paradigm of society, from the intersection of history to the dominance\\ of culture. A key clue in the future of art forms can be found in the continued decentralization of power in-\\to more hands. Being able to participate in social movements is also the standard for future artists to choo-\\se artists.The expression of works of art can be expressed in pictures, audio and video, or in any kind of be-\\havior. The key is whether you can use the tools you are good at, or be willing to expand your imagination \\to express your thinking about the encrypted world.}\\
        \specialrule{0em}{1.5pt}{1.5pt}
        \textbf{Gold} & \tabincell{l}{The outbreak of encryption has stirred up the entire art market, driven the reform and opening up of the art\\ field, broken the original concept, and brought new creative media. This is a challenge to the art industry\\ and a turning point, allowing creators to better integrate with the world. But artists still need standards, they\\ must have character, they must have the ability to innovate, and they must make their works clearer and mo-\\re expressive. Encryption starts from the bottom and has vigorous vitality. Everyone who participates in so-\\cial advancement is already participating in the creation of art.}\\
    \bottomrule[0.8pt]
    \end{tabular}
    \caption{Case studies on the joint summary (the first part) of the segment and the overall summary (the second part).}
    \label{tab:en_case_study}
\end{table*}

\section{Details of \textsc{ConvLM}} \label{sec:convlm}
\textsc{ConvLM} employs the architecture of large BART model. The training data is collected from Chinese social media, including Zhihu\footnote{\url{https://www.zhihu.com/}}, Weibo\footnote{\url{https://www.weibo.com/}} and Douban\footnote{\url{https://www.douban.com/}}. We totally crawl 10 billion pieces of dialogue data, and select 10 million pieces of high-quality data to train \textsc{ConvLM}. The pre-training objectives are speaker identification and response generation, wherein speaker identification is to predict the masked speaker indicators and response generation is to generate the response based on the context. We truncate the input sequence into 512. The batch size is set to 128. The learning rate is set to 5e-5 and linear updated along with the training process. We train the \textsc{ConvLM} for 300K steps, which takes around 50 hours on 16 NVIDIA A100 GPUs. We validate the model each 1K steps. We save the model with the best validation loss and the last 10 checkpoints.

\section{Screenshot of Annotation Platform}
\begin{figure*}[t!]
    \centering
    \includegraphics[width=1.0\linewidth]{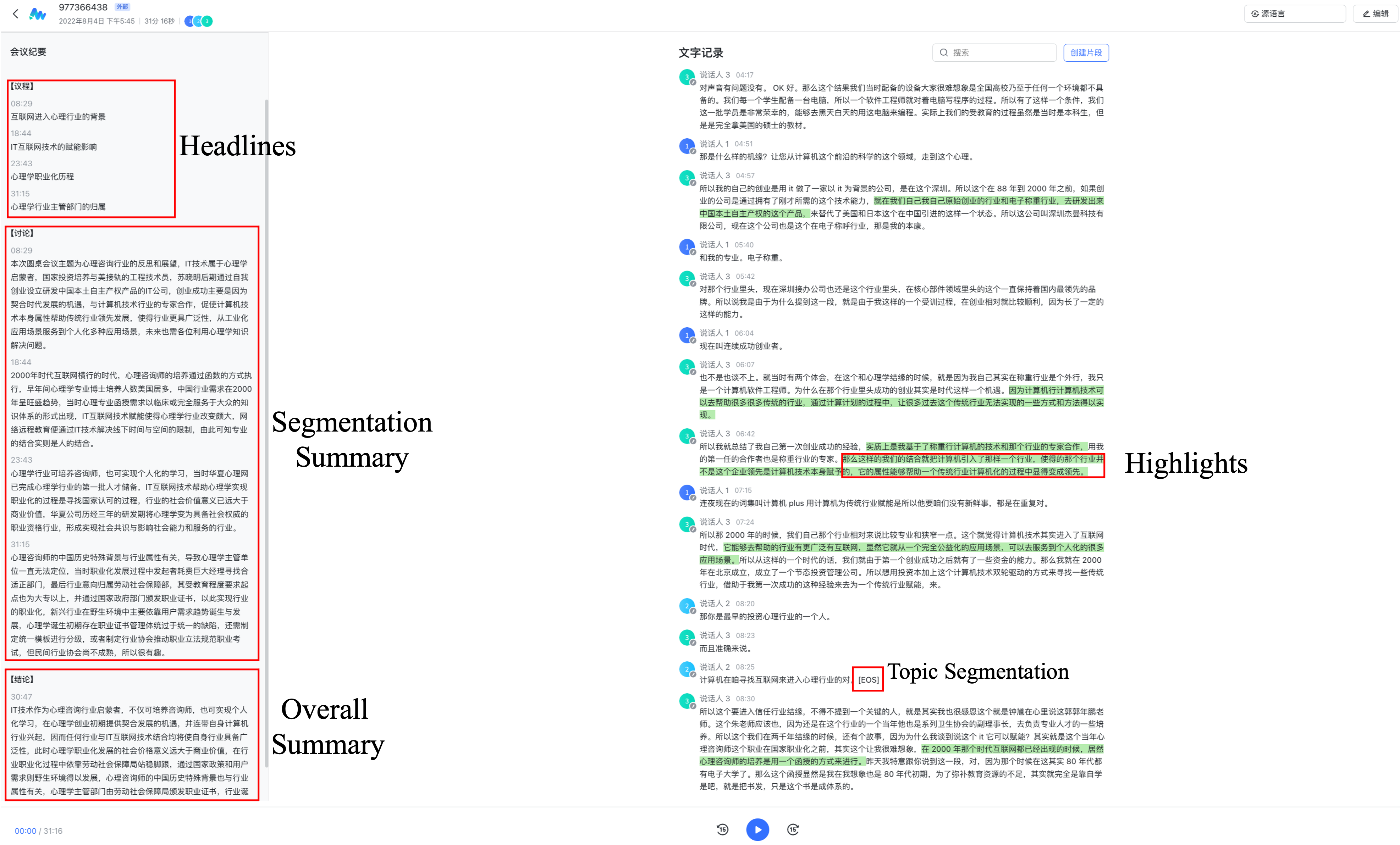}
    \caption{The screenshot of the annotation platform.}
    \label{fig:screenshot_anno}
\end{figure*}

See Figure \ref{fig:screenshot_anno}.

\end{CJK*}
\end{document}